\pdfoutput=1
\PassOptionsToPackage{table,dvipsnames}{xcolor}  

\documentclass[11pt]{article}

\usepackage{amsmath}

\usepackage[final]{acl}
\usepackage{booktabs}
\usepackage{hyperref}
\usepackage[utf8]{inputenc}
\usepackage{xcolor}
\usepackage[normalem]{ulem}
\definecolor{PrimaryBlue}{RGB}{0,0,255}
\definecolor{PrimaryRed}{RGB}{255,0,0}

\definecolor{TaskBG}{RGB}{235,240,255}

\definecolor{LightGray}{gray}{0.92}

\usepackage{times}
\usepackage{latexsym}
\usepackage{multirow}
\usepackage{tabularx}
\usepackage{pifont}
\usepackage{tcolorbox}
\usepackage[x11names]{xcolor}

\usepackage{multirow}      
\usepackage{booktabs}      
\usepackage{adjustbox}     
\usepackage{caption}       
\usepackage{geometry}      
\usepackage{amssymb}
\usepackage[T1]{fontenc}
\usepackage[utf8]{inputenc}
\usepackage{textcomp}
\usepackage{microtype}
\usepackage{inconsolata}
\usepackage{seqsplit}         

\usepackage{microtype}
\usepackage{booktabs}
\usepackage{amsfonts}
\usepackage{graphicx}
\usepackage{hyperref}
\usepackage{textcomp}
\usepackage{listings}
\usepackage{adjustbox}
\usepackage{xspace}    
\usepackage{multirow}
\usepackage{multicol}
\usepackage{xcolor} 
\usepackage{amsmath}
\usepackage{tcolorbox}
\usepackage[linesnumbered,boxed,ruled]{algorithm2e}
\usepackage{listings}

\newcommand\figref[1]{Fig.~\ref{#1}}

\newcommand\tabref[1]{Table~\ref{#1}}

\newcommand\secref[1]{\S\ref{#1}}

\newcommand\appref[1]{Appendix~\ref{#1}}
\newcommand{\fakeparagraph}[1]{\noindent\textbf{#1.}}

\usepackage{wasysym}
\usepackage{algorithmic}

\definecolor{codegreen}{rgb}{0,0.6,0}
\definecolor{codegray}{rgb}{0.5,0.5,0.5}
\definecolor{codepurple}{rgb}{0.58,0,0.82}
\definecolor{backcolour}{rgb}{0.95,0.95,0.92}

\lstdefinestyle{mystyle}{
  backgroundcolor=\color{backcolour},   commentstyle=\color{codegreen},
  keywordstyle=\color{magenta},
  numberstyle=\tiny\color{codegray},
  stringstyle=\color{codepurple},
  basicstyle=\ttfamily\footnotesize,
  breakatwhitespace=false,         
  breaklines=true,                 
  captionpos=b,                    
  keepspaces=true,                 
  numbers=left,                    
  numbersep=5pt,                  
  showspaces=false,                
  showstringspaces=false,
  showtabs=false,                  
  tabsize=2
}

\usepackage{balance} 
\usepackage{inconsolata} 
\usepackage{graphicx}
\usepackage{tabularx} 

\usepackage{xcolor}
\usepackage{graphicx}
\usepackage{framed}
\usepackage{pifont}
\definecolor{shadecolor}{rgb}{0.92,0.92,0.92}
\usepackage{tikz}
\usetikzlibrary{shapes.geometric, arrows.meta, positioning}

\tikzstyle{box} = [rectangle, rounded corners, minimum width=3.2cm, minimum height=1cm, text centered, draw=black, fill=gray!10]
\tikzstyle{process} = [rectangle, minimum width=3.2cm, minimum height=1cm, text centered, draw=black, fill=blue!10]
\tikzstyle{decision} = [diamond, draw=black, fill=yellow!30, minimum size=1.2cm, text centered, inner sep=0pt, aspect=2]
\tikzstyle{arrow} = [thick, ->, >=stealth]
\PassOptionsToPackage{table}{xcolor}

\title{Benchmarking Large Language Models Under Data Contamination: A Survey from Static to Dynamic Evaluation} 

\author{
  {\small \href{mailto:sc5687@columbia.edu}{Simin Chen}$^{1}$ \quad
  \href{mailto:yiming.chen@u.nus.edu}{Yiming Chen}$^{2}$\thanks{Corresponding authors: yiming.chen@u.nus.edu, zli536@ucr.edu} \quad
  \href{mailto:zli536@ucr.edu}{Zexin Li}$^{3}$\footnotemark[1] \quad
  \href{mailto:yifjia@isi.edu}{Yifan Jiang}$^{4}$ \quad
  \href{mailto:wan.512@osu.edu}{Zhongwei Wan}$^{5}$ \quad
  \href{mailto:heyixinn00@gmail.com}{Yixin He}$^{4}$} \\[0.4em]
  {\small \href{mailto:dezhiran@pku.edu.cn}{\textbf{Dezhi Ran}}$^{6}$ \quad
  \href{mailto:tianle.gu@mbzuai.ac.ae}{\textbf{Tianle Gu}}$^{7}$ \quad
  \href{mailto:haizhou.li@nus.edu.sg}{\textbf{Haizhou Li}}$^{2,8}$ \quad
  \href{mailto:taoxie@pku.edu.cn}{\textbf{Tao Xie}}$^{6}$ \quad
  \href{mailto:rayb@cs.columbia.edu}{\textbf{Baishakhi Ray}}$^{1}$} \\[0.6em]
  {\small $^{1}$Columbia University \quad
  $^{2}$National University of Singapore \quad
  $^{3}$University of California, Riverside} \\ 
  {\small $^{4}$University of Southern California \quad
  $^{5}$The Ohio State University \quad
  $^{6}$Peking University \quad} \\
  {\small 
  $^{7}$Tsinghua University \quad
  $^{8}$The Chinese University of Hong Kong, Shenzhen} \\[0.6em]
  {\href{https://github.com/SeekingDream/Static-to-Dynamic-LLMEval}{Static-to-Dynamic-LLMEval GitHub Repository}}
}

\usepackage{float}
\begin{document}

\maketitle

\begin{abstract}
In the era of evaluating large language models (LLMs), data contamination has become an increasingly prominent concern. To address this data contamination risk, LLM benchmarking has evolved from a \textit{static} to a \textit{dynamic} paradigm.
In this work, we conduct an in-depth analysis of existing \textit{static} and \textit{dynamic} benchmarks for evaluating LLMs. 
We first examine methods that enhance \textit{static} benchmarks and identify their inherent limitations. We then highlight a critical gap—the lack of standardized criteria for evaluating \textit{dynamic} benchmarks. Based on this observation, we propose a series of optimal design principles for \textit{dynamic} benchmarking and analyze the limitations of existing \textit{dynamic} benchmarks.  
This survey provides a concise yet comprehensive overview of recent advancements in data contamination research, offering valuable insights and a clear guide for future research efforts. 
We maintain a GitHub repository to continuously collect both static and dynamic benchmarks for LLMs.

\end{abstract}








\section{Introduction}

The field of natural language processing (NLP) has advanced rapidly in recent years, driven by breakthroughs in Large Language Models (LLMs) such as GPT-4, Claude3, and DeepSeek~\cite{achiam2023gpt, liu2024deepseek, wan2023efficient}. Trained on vast amounts of Internet-sourced data, these models have demonstrated remarkable capabilities across various applications, including code generation, text summarization, computer use, and mathematical reasoning~\cite{codeforces, ran2024guardian,hu-etal-2024-reasoning}.

To develop and improve LLMs, beyond advancements in model architectures and training algorithms, a crucial area of research focuses on effectively evaluating their intelligence. Traditionally, LLM evaluation has relied on \textit{static} benchmarking, which involves using carefully curated human-crafted datasets and assessing model performance with appropriate metrics~\cite{wang-etal-2018-glue, achiam2023gpt, gunasekar2023textbooks,2025arXiv250102863R}.

However, because these \textit{static} benchmarks are released on the Internet for transparent evaluation, and LLMs gather as much data as possible from the Internet for training, potential data contamination is unavoidable~\cite{magar-schwartz-2022-data, deng-etal-2024-investigating, li-etal-2024-open-source, balloccu-etal-2024-leak}. 
Data contamination occurs when benchmark data is inadvertently included in the training phase of LLMs, leading to inflated and misleading performance assessments. Although this issue has long been recognized—rooted in the fundamental machine learning principle of separating training and test sets—it has become more critical with the rise of LLMs, which often scrape vast amounts of publicly available Internet data~\cite{achiam2023gpt}, increasing the risk of contamination. 
Furthermore, due to privacy and commercial concerns, tracing the exact training data of these models is challenging—if not impossible—complicating efforts to detect and mitigate potential contamination.

\begin{figure}
    \centering
    \includegraphics[width=0.46\textwidth]{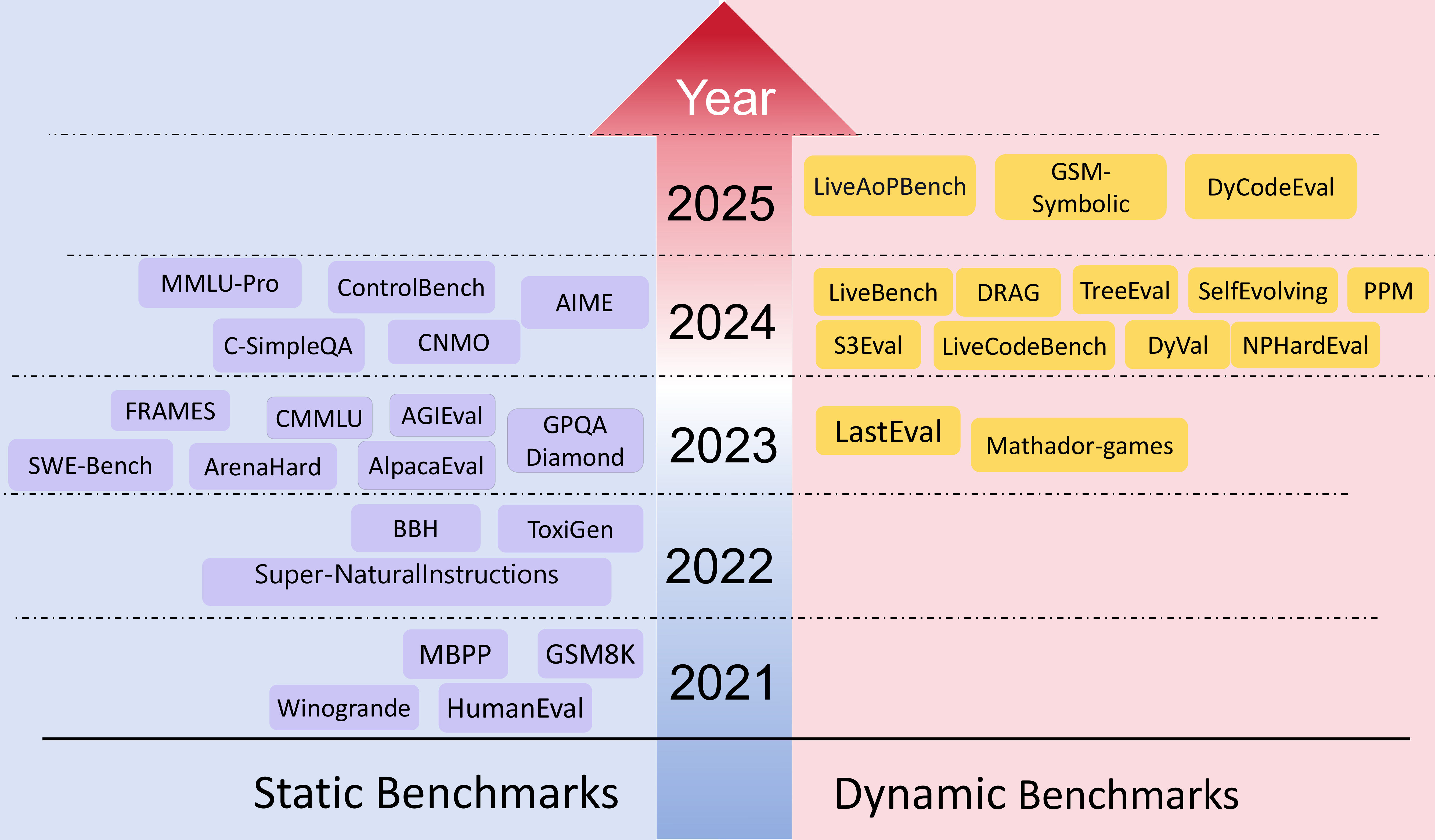}
    \caption{The progress of benchmarking LLMs.}
    \label{fig:transition}
\end{figure}

To mitigate the risk of data contamination in LLM benchmarking, researchers have proposed several enhancements to static evaluation methods, including data encryption~\cite{jacovi-etal-2023-stop} and post-hoc contamination detection~\cite{shidetecting}. However, due to the inherent limitations of static methods—such as unverifiable data exposure—these enhancements have seen limited adoption. As a result, researchers have shifted toward new \textit{dynamic} benchmarking paradigms, as illustrated in \figref{fig:transition}. Dynamic methods aim to reduce contamination risk either by continuously updating benchmark datasets based on LLM training timestamps~\cite{white2024livebench, jain2024livecodebenchholisticcontaminationfree}, or by regenerating test data to reconstruct and replace original benchmarks~\cite{chen2024ppm, zhou2025gsm, mirzadeh2025gsmsymbolic}.

Although many dynamic benchmarking methods have been proposed to promote fair and transparent evaluation of LLMs, most existing work primarily highlights the advantages of these dynamic benchmarks~\cite{white2024livebench}. However, the question remains: \textit{what are the potential trade-offs of using dynamic benchmarks to evaluate LLMs?}
The limitations of dynamic benchmarking—such as the computational overhead of continuous updates, and the need for reliable timestamp metadata—are not yet fully explored.

Moreover, existing surveys on LLM data contamination have mainly focused on post-hoc detection methods~\cite{deng-etal-2024-unveiling, ravaut2024much, xu2024benchmark, dong-etal-2024-generalization, balloccu-etal-2024-leak}, offering little attention to the emerging landscape of dynamic benchmarking strategies. Considering the growing importance and adoption of dynamic benchmarking methods, it is essential to assess their effectiveness and limitations. Unfortunately, our empirical survey of existing dynamic benchmarking methods reveals that their evaluations are highly fragmented. To date, there is no systematic work that defines clear evaluation criteria for dynamic benchmarks themselves. Moreover, existing reviews often overlook a detailed comparison of the strengths and weaknesses of different dynamic methods, leaving a gap in understanding their practical trade-offs and applicability.

To bridge this gap, we first conduct a systematic survey of benchmarking methods for LLMs designed to mitigate the risk of data contamination, covering both \textit{static} and \textit{dynamic} benchmarks. We summarize state-of-the-art methods and provide an in-depth discussion of their strengths and limitations.   Furthermore, we are the first to summarize and abstract a set of criteria for evaluating \textit{dynamic} benchmarks. Our study reveals that existing \textit{dynamic} benchmarks do not fully satisfy these proposed criteria, implying the imperfection of current design. 
We hope that our criteria will provide valuable insights for the future design and standardization of \textit{dynamic} benchmarking methods.

\begin{figure*}
    \centering
    \includegraphics[width=0.88\textwidth]{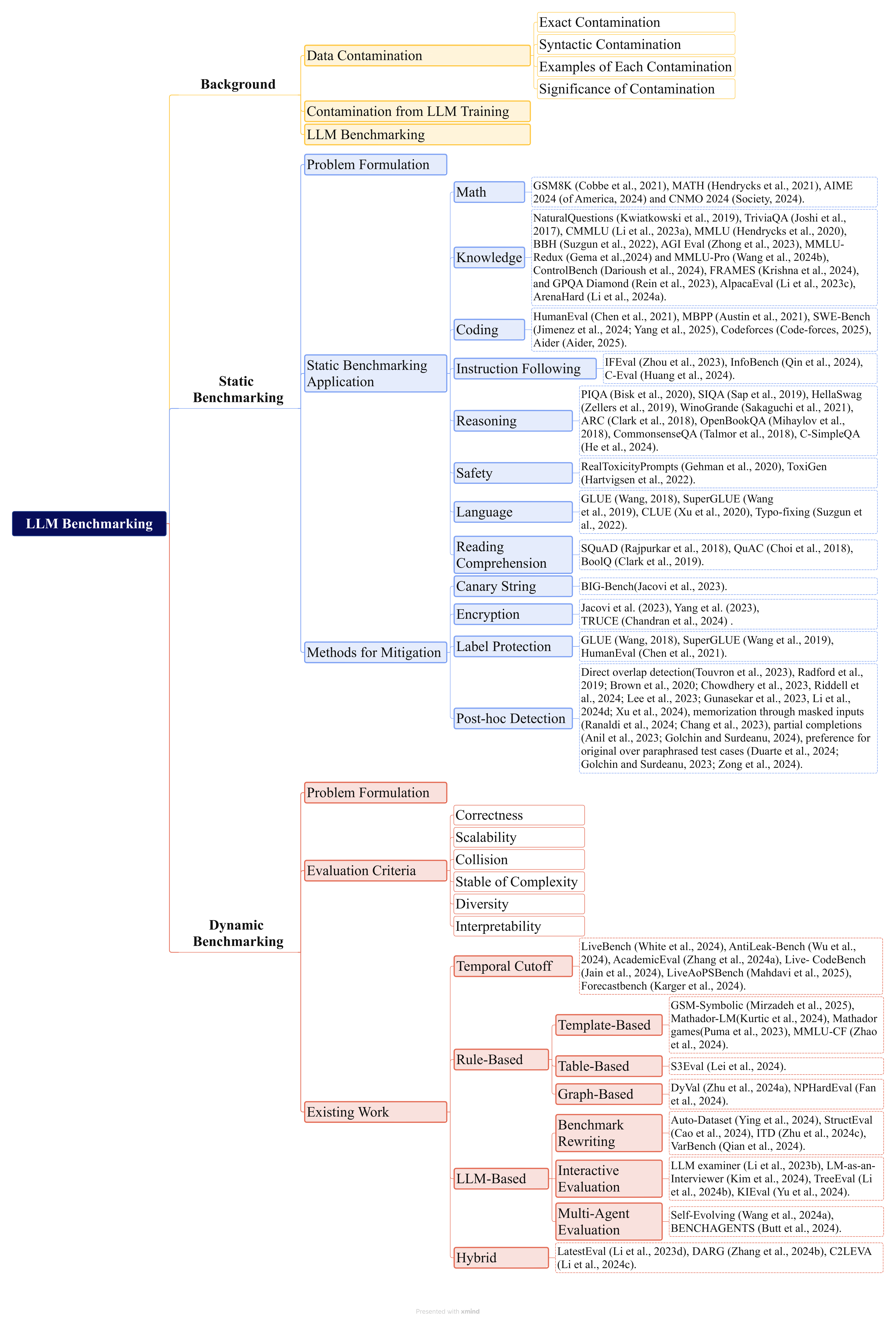}
    \caption{ Taxonomy of research on benchmarking LLMs.}
    \label{fig:Taxonomy}
\end{figure*}

The paper is organized as shown in \figref{fig:Taxonomy}. We first review the background on data contamination (\secref{sec:background}), and then survey \textit{static} benchmarks and their enhancements for mitigating data contamination (\secref{sec:static}). Next, we introduce key principles and existing methods for \textit{dynamic} benchmarking (\secref{sec:dynamic}). Finally, we discuss open challenges and future directions (\secref{sec:discussion}).


\begin{table*}[htbp]
  \centering

  \renewcommand{\arraystretch}{1.2} 
  
  \resizebox{0.8\textwidth}{!}{
  \begin{tabular}{p{4cm} p{6cm} p{10cm}}
    \toprule
    \toprule
    \textbf{Contamination Type} & \textbf{Training Data} & \textbf{Testing Data} \\
    \hline
    Exact Contamination & 
    \texttt{Write a Python function to check if a number is prime.} & 
    \texttt{Write a Python function to check if a number is prime.} \\
    \hline
    Syntactic Contamination & 
    \texttt{Write a Python function to check if a number is prime.} & 
    \texttt{You are a helpful code assistant for Python. Write a Python function to check if a number is prime.} \\
    \bottomrule
    \bottomrule
  \end{tabular}
  }
  \caption{Examples of data contamination in LLMs.}  
  \label{tab:contamination_examples}
\end{table*}

\section{Background}
\label{sec:background}

\subsection{Data Contamination}

Data contamination arises when LLM training data $\mathcal{D}_{\text{train}}$ improperly overlaps with evaluation data $\mathcal{D}_{\text{test}}$, undermining performance validity. We review existing work and formalize the definition.

\fakeparagraph{Exact Contamination} Exact contamination occurs when there is any exact duplicate in the benchmark dataset 
\[
\exists \, d \quad \text{s.t.} \quad d \in \mathcal{D}_{\text{train}} \text{ and } d \in \mathcal{D}_{\text{test}}
\]
In other words, there exists a data point $d$ that is in both $\mathcal{D}_{\text{train}}$ and $\mathcal{D}_{\text{test}}$.
Common cases include verbatim test examples appearing in training corpora, code snippets from benchmark implementations, or documentation leaks.

\fakeparagraph{Syntactic Contamination} Syntactic contamination occurs when a test data point could be found in the training dataset after  a syntactic transformation, such that
\[
\exists \, d \quad \text{s.t.} \quad \mathcal{F}_{\text{syntactic}}(d) \in \mathcal{D}_{\text{train}} \text{ and } d \in \mathcal{D}_{\text{test}}
\]
where \(\mathcal{F}_{\text{syntactic}}\) denotes syntactic transformations like punctuation normalization, whitespace modification, synonym substitution, morphological variations, or syntactic paraphrasing while preserving lexical meaning.

\fakeparagraph{Examples of Each Contamination}
We provide contamination examples in \tabref{tab:contamination_examples}. 
Syntactic contamination occurs when test data is rephrased from training data using a prefix. Whether such syntactic contamination constitutes true contamination is debated, as it is difficult to separate memorization from reasoning. In this work, we treat such transformations as contamination, since some NLP tasks rely heavily on syntax.

\fakeparagraph{Significance of Contamination}
Data contamination poses a serious threat to the integrity of LLM benchmarking, particularly as models grow in scale and are trained on vast publicly available corpora. Without proper safeguards, evaluations may inadvertently test models on data that they have seen during training, leading to inflated performance metrics and misleading claims about generalization and robustness. Recent studies underscore this concern: \citet{schaeffer2023pretraining} demonstrates that pretraining on test data can significantly distort evaluation outcomes; \citet{balloccu-etal-2024-leak} reveal how easily data contamination and evaluation malpractices can occur in closed-source LLMs; \citet{xu2024benchmarking} propose methods to quantify such contamination; and \citet{deng-etal-2024-unveiling} provide a comprehensive survey of existing risks and mitigation strategies.
The issue gained public attention when Meta’s LLaMA 4 faced allegations of using a non-public version fine-tuned for benchmark gains~\cite{bab407_llama4_2024}, raising concerns about evaluation transparency—despite Meta’s denial of test set exposure. Such cases underscore the need for contamination-aware benchmarking to accurately assess LLM performance on truly unseen data.
We also present a proof-of-concept evaluation in \appref{sec:appendix_demo} to highlight the impact of data contamination.

\subsection{Contamination Source}
Data contamination can occur during the pre-training, post-training, or fine-tuning phases of LLM development.
Unlike traditional models with clear separations between training and evaluation data, LLMs are pre-trained on massive, diverse datasets—often scraped from the web (e.g., FineWeb~\cite{penedo2024the}), increasing the risk of evaluation data overlap. In the post-training phase, models are further fine-tuned on large human-annotated~\cite{mukherjee2023orca,kim-etal-2023-cot} or synthetic datasets~\cite{ding-etal-2023-enhancing,OpenHermes25,wang2023openchat} that may resemble evaluation tasks, further compounding contamination risks. Although retrieval-based detection methods~\cite{team2024gemini,achiam2023gpt} exist, the sheer scale and complexity of training corpora make it difficult to entirely exclude evaluation data. Additionally, many LLMs keep their training data proprietary~\cite{dubey2024llama,yang2024qwen2}, complicating the accurate assessment of their true performance and highlighting the need for fair and reliable benchmarks. This opacity further exacerbates data contamination, as it impedes the community’s ability to verify and mitigate potential overlaps between training and evaluation data.

\subsection{LLM Benchmarking}


As LLMs evolve into general-purpose task solvers, it is crucial to develop benchmarks that provide a holistic view of their performance.
To this end, significant human effort has been dedicated to building comprehensive benchmarks that assess various aspects of model performance. For example, instruction-following tasks evaluate a model's ability to interpret and execute commands~\citep{zhou2023instruction,qin-etal-2024-infobench,huang2024c}, while coding tasks assess its capability to generate and understand programming code~\citep{chen2021evaluating,austin2021program,jimenez2024swebench,codeforces,aider}.
Despite their usefulness, static benchmarks face challenges as LLMs evolve rapidly and continue training on all available data~\citep{villalobos2022will}. Over time, unchanging benchmarks may become too easy for stronger LLMs or introduce data contamination issues.
Recognizing this critical problem, contamination detectors have been developed to quantify contamination risks, and dynamic benchmarks have been proposed to mitigate these issues.

\section{Static Benchmarking}
\label{sec:static}

\subsection{Problem Formulation}
A static benchmark is given by
$ \mathcal{D} = (\mathcal{X}, \mathcal{Y}, \mathcal{S}(.)) $, 
where \( \mathcal{D} \) represents the seed dataset, consisting of input prompts \( \mathcal{X} \), expected outputs \( \mathcal{Y} \), and a scoring function \( \mathcal{S}(\cdot) \) that evaluates the quality of an LLM's outputs by comparing them against \( \mathcal{Y} \). 

\subsection{Static Benchmark Applications}

\fakeparagraph{Math}
Math benchmarks evaluate a model’s ability to solve multi-step math problems. 
Datasets such as GSM8K~\citep{cobbe2021training} and MATH~\citep{hendrycks2021measuring} require models to work through complex problems. Recent challenges like AIME 2024~\citep{maa2024aime} and CNMO 2024~\citep{cnmo2024} further test a model’s capacity to tackle diverse and intricate math tasks.

\fakeparagraph{Coding}
Coding benchmarks measure a model’s ability to generate and debug code. HumanEval~\citep{chen2021evaluating} and MBPP~\citep{austin2021program} test code synthesis and debugging, whereas SWE-Bench~\citep{jimenez2024swebench,yang2024swebenchmultimodal} addresses more advanced challenges. Competitive platforms like Codeforces~\citep{codeforces} and datasets like Aider~\citep{aider} further probe dynamic problem solving.
There are also benchmarks that evaluate LLMs’ ability to judge the correctness of code~\citep{jiang2025codejudgebench}.

\fakeparagraph{Instruction Following}
Instruction benchmarks evaluate a model’s ability to comprehend and execute detailed directives. Datasets like IFEval~\citep{zhou2023instruction} and InfoBench~\citep{qin-etal-2024-infobench} simulate real-world scenarios requiring clear, step-by-step guidance, with C-Eval~\citep{huang2024c} focusing on Chinese instructions.

\fakeparagraph{Other Applications}
We provide a detailed introduction to other applications in \appref{sec:appendix_applications}, along with a further analysis on enhancing static benchmarks in Appendix~\ref{sec:enhancement}.

\section{Dynamic Benchmarking}
\label{sec:dynamic}

\subsection{Problem Formulation}

A dynamic benchmark is defined as  
$
\mathcal{B}_{\text{dynamic}} = (\mathcal{D}, T(\cdot)), \quad 
\mathcal{D} = (\mathcal{X}, \mathcal{Y}, \mathcal{S}(\cdot))
$
where \( \mathcal{D} \) represents the static benchmark dataset. 
The transformation function \( T(\cdot) \) modifies the dataset during the benchmarking to avoid possible data contamination.
The dynamic dataset for the evaluation of an LLM can then be expressed as
$
        \mathcal{D}_t = T_t(\mathcal{D}),  \quad
        \forall t \in \{1, \dots, N\}
$
where \( \mathcal{D}_t \) represents the evaluation dataset at the timestamp \( t \), and \( N\) is the total timestamp number, which could be finite or infinite. 
If the seed dataset $\mathcal{D}$ is empty, the dynamic benchmarking dataset will be created from scratch.

\subsection{Criteria Summarization and Abstraction} 

While many dynamic benchmarking methods have been proposed to evaluate LLMs, the criteria for evaluating these benchmarks themselves remain non-standardized. To address this gap, we analyze existing evaluation practices and abstract them into a unified framework.
We review over 50 dynamic benchmarking papers, focusing specifically on how they evaluate their own benchmarks. Although many of these works include some form of self-evaluation, the dynamic benchmarking methods are often incomplete, or lack depth. For example, DyVal2 evaluates benchmark complexity and correctness, but does not address the interpretability of the benchmark construction process.

To systematize this landscape, we identify a unified set of evaluation criteria and present them in \tabref{tab:dynamic}. We then assess whether each dynamic benchmark fully supports, partially supports, or does not support each criterion.
For instance, in the case of correctness: benchmarks with built-in guarantees—such as those using temporal cutoffs or rule-based generation—are marked as ``supported''. Benchmarks generated using LLMs are marked as ``partially supported'' if they include validation (e.g., human or automated checks); otherwise, they are labeled ``not supported''. More guidance for classifying each dynamic benchmark could be found in \appref{sec:appendix_labeling}.

\begin{table*}[htbp]
  \centering
   
  \resizebox{0.88\textwidth}{!}{
    \begin{tabular}{clcccccc}
    \toprule
    \toprule
    \multirow{2}[2]{*}{\textbf{Dynamic Mechanisms}} & \multirow{2}[2]{*}{\textbf{Benchmark Name}} & \multicolumn{6}{c}{\textbf{Evaluation Criteria}} \\
          &       & \textbf{Correctness} & \textbf{Scalability} & \textbf{Collision} & \textbf{Stability of Complexity} & \textbf{Diversity} & \textbf{Interpretability} \\
    \midrule
    \multirow{6}[2]{*}{\textbf{Temporal Cutoff}} & LiveBench~\citep{white2024livebench} & \CIRCLE  & \LEFTcircle & \LEFTcircle & \Circle & \Circle & \CIRCLE  \\
          & AcademicEval~\citep{zhang2024academiceval}  & \CIRCLE  & \LEFTcircle & \LEFTcircle & \Circle & \Circle & \CIRCLE  \\
          & LiveCodeBench~\citep{jain2024livecodebenchholisticcontaminationfree} & \CIRCLE  & \LEFTcircle & \LEFTcircle & \Circle & \Circle & \CIRCLE  \\
          & LiveAoPSBench~\citep{mahdavi2025leveraging} & \CIRCLE  & \LEFTcircle & \LEFTcircle & \Circle & \Circle & \CIRCLE  \\
          & AntiLeak-Bench~\citep{wu2024antileak} & \CIRCLE  & \CIRCLE  & \LEFTcircle & \CIRCLE  & \Circle & \CIRCLE  \\
    \midrule
    \multirow{6}[2]{*}{\textbf{Rule-Based}} & S3Eval~\citep{lei-etal-2024-s3eval} & \CIRCLE  & \CIRCLE  & \LEFTcircle & \CIRCLE  & \LEFTcircle & \CIRCLE  \\
          & DyVal~\citep{zhu2024dyval}  & \CIRCLE  & \CIRCLE  & \CIRCLE  & \CIRCLE  & \LEFTcircle & \CIRCLE  \\
        & MMLU-CF~\citep{zhao2024mmlu} & \CIRCLE  & \LEFTcircle & \Circle & \CIRCLE  & \LEFTcircle & \LEFTcircle \\
          & NPHardEval~\citep{fan-etal-2024-nphardeval} & \CIRCLE  & \CIRCLE  & \CIRCLE  & \CIRCLE  & \LEFTcircle & \CIRCLE  \\
          & GSM-Symbolic~\citep{mirzadeh2025gsmsymbolic} & \CIRCLE  & \Circle & \LEFTcircle & \CIRCLE  & \LEFTcircle & \CIRCLE  \\
          & PPM~\citep{chen2024ppm}   & \CIRCLE  & \CIRCLE  & \CIRCLE  & \LEFTcircle & \LEFTcircle & \CIRCLE  \\
          & GSM-Infinite~\citep{zhou2025gsm} & \CIRCLE  & \CIRCLE  & \CIRCLE  & \CIRCLE  & \LEFTcircle & \CIRCLE  \\
    \midrule
    \multirow{7}[2]{*}{\textbf{LLM-Based}} & Auto-Dataset~\citep{ying2024automating} & \LEFTcircle & \CIRCLE  & \LEFTcircle & \CIRCLE  & \CIRCLE  & \Circle \\
          & LLM-as-an-Interviewer~\citep{kim2024llm} & \LEFTcircle & \CIRCLE  & \LEFTcircle & \LEFTcircle & \CIRCLE  & \Circle \\
          & TreeEval~\citep{li2024treeeval} & \LEFTcircle & \CIRCLE  & \LEFTcircle & \LEFTcircle & \LEFTcircle & \Circle \\
          & BeyondStatic~\citep{li2023beyond} & \LEFTcircle & \CIRCLE  & \LEFTcircle & \Circle & \CIRCLE  & \Circle \\
          & StructEval~\citep{cao-etal-2024-structeval}  & \LEFTcircle & \CIRCLE  & \CIRCLE  & \CIRCLE  & \CIRCLE  & \Circle \\
          & Dynabench~\citep{kiela-etal-2021-dynabench}  & \LEFTcircle & \LEFTcircle & \LEFTcircle & \LEFTcircle & \CIRCLE  & \Circle \\
          & Self-Evolving~\citep{wang2024selfevolving} & \LEFTcircle & \CIRCLE  & \LEFTcircle & \CIRCLE  & \CIRCLE  & \Circle \\
    \midrule
    \multirow{3}[2]{*}{\textbf{Hybrid }} & DARG~\citep{zhang2024darg}  & \LEFTcircle & \CIRCLE  & \LEFTcircle & \CIRCLE  & \CIRCLE  & \LEFTcircle \\
    & LatestEval~\citep{li2023avoiding} & \CIRCLE  & \LEFTcircle & \LEFTcircle & \Circle & \Circle & \CIRCLE  \\
          & C2LEVA~\citep{li-etal-2025-c2leva} & \LEFTcircle & \CIRCLE  & \LEFTcircle & \LEFTcircle & \CIRCLE  & \LEFTcircle \\
    \bottomrule
    \bottomrule
    \end{tabular}%
    
    }
    \caption{Existing dynamic benchmarks and their quality on our summarized criteria. $\CIRCLE$ represents support, $\LEFTcircle$ represents partial support, and $\Circle$ represents no support. }   
  \label{tab:dynamic}%
\end{table*}%

\subsection{Summarized Evaluation Criteria}

\subsubsection{Correctness}
The first criterion for evaluating the quality of dynamic benchmarking is \textsf{Correctness}. If the correctness of the generated dataset cannot be guaranteed, the benchmark may provide a false sense of reliability when applied to benchmarking LLMs, leading to misleading evaluations.  
We quantify the correctness of dynamic benchmarks as
$$
    \textsf{Correctness} = \mathbb{E}_{i=1}^{N}  
    \mathcal{S} \big( \mathcal{Y}_i, \mathcal{G}(\mathcal{X}_i) \big)
$$
where \( \mathcal{X}_i \) and \( \mathcal{Y}_i \) represent the input and output of the \( i^{th} \) transformation, respectively. The function \( \mathcal{G}(\cdot) \) is an oracle that returns the ground truth  of its input, ensuring an objective reference for correctness evaluation. For example, the function \( \mathcal{G}(\cdot) \) could be a domain-specific annotator.  
This equation can be interpreted as the expected alignment between the outputs of the transformed data set and their corresponding ground truth values, measured using the scoring function \( \mathcal{S}(\cdot) \). A higher correctness score indicates that the dynamic benchmark maintains correctness to the ground truth.

\subsubsection{Scalability}
The next evaluation criterion is scalability, which measures the ability of dynamic benchmarking methods to generate large-scale benchmark datasets. A smaller dataset can introduce more statistical errors during the benchmarking process. Therefore, an optimal dynamic benchmark should generate a larger dataset while minimizing associated costs. The scalability of a dynamic benchmark is quantified as
$$
    \textsf{Scalability} = \mathbb{E}_{i=1}^{N} \left[ \frac{\lVert T_i(\mathcal{D}) \rVert}{\lVert \mathcal{D} \rVert \times \textsf{Cost}(T_i)} \right]
$$
This equation represents the expectation over the entire transformation space, where \( \lVert T_i(\mathcal{D}) \rVert \) is the size of the transformed dataset, and \( \lVert \mathcal{D} \rVert \) is the size of the original dataset. The function \( \textsf{Cost}(\cdot) \) measures the cost associated with the transformation process, which could include monetary cost, time spent, or manual effort according to the detailed scenarios.
This equation could be interpreted as the proportion of data that can be generated per unit cost.

\subsubsection{Collision}

One of the main motivations for dynamic benchmarking is to address the challenge of balancing transparent benchmarking with the risk of data contamination. Since the benchmarking algorithm is publicly available, an important concern arises:  \textit{If these benchmarks are used to train LLMs, can they still reliably reflect the true capabilities of the LLMs?}
To evaluate the robustness of a dynamic benchmark against this challenge, we introduce the concept of \textit{collision} in dynamic benchmarking. Collision refers to the extent to which different transformations of the benchmark dataset produce overlapping data, potentially limiting the benchmark’s ability to generate novel and diverse test cases. To quantify collision, we propose the following metrics

\[
\begin{split}
    & \textsf{Collision Rate} = \mathbb{E}_{\substack{i, j = 1, \, i \neq j}}^{N}  
    \left[ \frac{\lVert \mathcal{D}_i \cap \mathcal{D}_j \rVert }{ \lVert \mathcal{D} \rVert} \right]\\
    & \textsf{Repeat} = \mathbb{E}_{i=1}^{N} \left[ k \mid k = \min \left\{ \bigcup_{j=1}^{k} \mathcal{D}_j \supseteq \mathcal{D}_i \right\} \right]
\end{split}
\]  
\textsf{Collision Rate} measures the percentage of overlap between two independently transformed versions of the benchmark dataset, indicating how much potential contamination among two trials. \textsf{Repeat Trials} quantifies the expected number of transformation trials required to fully regenerate an existing transformed dataset \( T_i(\mathcal{D}) \), providing insight into the benchmark’s ability to produce novel variations.  
These metrics help assess whether a dynamic benchmark remains effective in evaluating LLM capabilities, even when exposed to potential training data contamination.

\subsubsection{Stability of Complexity}

Dynamic benchmarks must also account for complexity to help users determine whether a performance drop in an LLM on the transformed dataset is due to potential data contamination or an increase in task complexity. If a dynamic transformation increases the complexity of the seed dataset, a performance drop is expected, even without data contamination. However, accurately measuring the complexity of a benchmark dataset remains a challenging task. Existing work has proposed various complexity metrics, but these are often domain-specific and do not generalize well across different applications. For example, DyVal~\citep{zhu2024dyval} proposes applying graph complexity to evaluate the complexity of reasoning problems.
Formally, given a complexity measurement function \( \Psi(\cdot) \), the stability can be formulated as 
\[
    \textsf{Stability} = \text{Var}(\Psi(D_i))
\]  
This equation can be interpreted as the variance in complexity across different trials, where high variance indicates that the dynamic benchmarking method is not stable.  


\subsubsection{Diversity }
The diversity metric can be categorized into two components: \textsf{external diversity} and \textsf{internal diversity}.  External diversity measures the variation between the transformed dataset and the seed dataset. Internal diversity quantifies the differences between two transformation trials.  
\[
    \begin{aligned}
        \textsf{External Diversity} &= \mathbb{E}_{i = 1}^{N} \Theta(\mathcal{D}_i, \mathcal{D}) \\
        \textsf{Internal Diversity} &= \mathbb{E}_{\substack{i, j = 1,  i \neq j}}^{N} \Theta(\mathcal{D}_i, \mathcal{D}_j)
    \end{aligned}
\]
where \( \Theta(\cdot) \) is a function that measures the diversity between two datasets. For example, it could be the N-gram metrics or the reference-based metrics, such as BLEU scores.

\subsubsection{Interpretability} 
Dynamic benchmarking generates large volumes of transformed data, making manual verification costly and challenging. To ensure correctness, the transformation process must be interpretable. Interpretable transformations reduce the need for extensive manual validation, lowering costs. Rule-based or manually crafted transformations are inherently interpretable, while LLM-assisted transformations depend on the model's transparency and traceability. In such cases, additional mechanisms like explainability tools or human-in-the-loop validation may be needed to ensure reliability and correctness.



\subsection{Existing Work}

Table~\ref{tab:static-benchmarks} summarizes recent dynamic benchmarks.
Dynamic benchmarking methods can be categorized into four types: temporal cutoff, rule-based generation, LLM-based generation, and hybrid.

\subsubsection{Temporal Cutoff} 
Since LLMs typically have a knowledge cutoff date, using data collected after this cutoff to construct datasets can help evaluate the model while mitigating data contamination. 
This type of method has been widely adopted to construct reliable benchmarks that prevent contamination~\cite{uddin2024unseentimeqa}.
LiveBench~\citep{white2024livebench} collects questions based on the latest information source, e.g., math competitions from the past 12 months, with new questions added and updated every few months.
AntiLeak-Bench~\citep{wu2024antileak} generates queries about newly emerged knowledge that was unknown before the model's knowledge cutoff date to eliminate potential data contamination.
AcademicEval~\citep{zhang2024academiceval} designs academic writing tasks on latest arXiv papers.
LiveCodeBench~\citep{jain2024livecodebenchholisticcontaminationfree} continuously collects new human-written coding problems from online coding competition platforms like LeetCode.
LiveAoPSBench~\citep{mahdavi2025leveraging} collects live math problems from the Art of Problem Solving forum.
Forecastbench~\citep{smith2024forecastbench} updates new forecasting questions on a daily basis from different data sources, e.g., prediction markets.

\fakeparagraph{Limitations}
The collection process typically requires significant human effort~\citep{white2024livebench,jain2024livecodebenchholisticcontaminationfree}, and continuous updates demand ongoing human involvement. Despite the popularity of temporal cutoffs, using recent information from competitions to evaluate LLMs can still lead to data contamination, as these problems are likely to be reused in future competitions~\citep{wu2024antileak}. Verification is often overlooked in these live benchmarks~\citep{white2024livebench}.

\subsubsection{Rule-Based Generation}
The method of rule-based generation synthesizes new test cases based on predefined rules, featuring an extremely low collision probability~\citep{zhu2024dyval}.

\fakeparagraph{Template-Based}
GSM-Symbolic~\citep{mirzadeh2025gsmsymbolic} creates dynamic math benchmarks by using query templates with placeholder variables, which are randomly filled to generate diverse problem instances.
Mathador-LM~\citep{kurtic-etal-2024-mathador} generates evaluation queries by adhering to the rules of Mathador games~\citep{PUMA2023105587} and varying input numbers.
MMLU-CF~\citep{zhao2024mmlu} follows the template of multiple-choice questions and generates novel samples by shuffling answer choices and randomly replacing incorrect options with ``None of the other choices''.

\fakeparagraph{Table-Based}
S3Eval~\citep{lei-etal-2024-s3eval} evaluates the reasoning ability of LLMs by assessing their accuracy in executing random SQL queries on randomly generated SQL tables.

\fakeparagraph{Graph-Based}
In this category, LLMs are evaluated with randomly generated graphs.
For instance, DyVal~\citep{zhu2024dyval} assesses the reasoning capabilities of LLMs using randomly generated directed acyclic graphs (DAGs). 
The framework first constructs DAGs with varying numbers of nodes and edges to control task difficulty. 
For example, in arithmetic reasoning tasks, leaf nodes represent random numeric values, while edges correspond to randomly assigned arithmetic operators. 
These DAGs are then transformed into natural language descriptions through rule-based conversion. Finally, the LLM is evaluated by querying it for the value of the root node.
Similarly, NPHardEval~\citep{fan-etal-2024-nphardeval} evaluates the reasoning ability of LLMs on well-known P and NP problems, such as the Traveling Salesman Problem (TSP). 
Random graphs of varying sizes are synthesized as inputs for TSP to assess the LLM's performance.
\citet{xie2024memorization} automatically construct Knights and Knaves puzzles with random reasoning graph.

\fakeparagraph{Limitations} 
The pre-defined rules may limit sample diversity, and publicly available rule-generated data may increase the risk of in-distribution contamination during training~\citep{tu2024dice}.

\subsubsection{LLM-Based Generation}

\fakeparagraph{Benchmark Rewriting}
In this category, LLMs are employed to rewrite samples from existing static benchmarks, which may be contaminated.
Auto-Dataset~\citep{ying2024automating} prompts LLMs to generate two types of new samples: one that retains the stylistics and essential knowledge of the original, and the other that presents related questions at different cognitive levels~\citep{bloom1956handbook}.
StructEval~\citep{cao-etal-2024-structeval} expands on examined concepts from the original benchmark by using LLMs and knowledge graphs to develop a series of extended questions.
ITD~\citep{zhu-etal-2024-inference} utilizes a contamination detector~\citep{shidetecting} to identify contaminated samples in static benchmarks and then prompts an LLM to rewrite them while preserving their difficulty levels.
VarBench~\citep{qian-etal-2024-varbench} prompts LLMs to generate new ones.

\fakeparagraph{Interactive Evaluation} 
In this category, inspired by the human interview process, LLMs are evaluated through multi-round interactions with an LLM ~\citep{li2023beyond}.
LLM-as-an-Interviewer~\citep{kim2024llm} employs an interviewer LLM that first paraphrases queries from existing static benchmarks and then conducts a multi-turn evaluation by posing follow-up questions or providing feedback on the examined LLM's responses.
TreeEval~\citep{li2024treeeval} begins by generating an initial question on a given topic using an LLM. Based on the previous topic and the examined LLM’s response, it then generates follow-up subtopics and corresponding questions to further assess the model.
KIEval~\citep{yu-etal-2024-kieval} generates follow-up questions based on the evaluated model's response to an initial question from a static benchmark.

\fakeparagraph{Multi-Agent Evaluation} 
Inspired by the recent success of multi-agent systems~\citep{guo2024large}, multi-agent collaborations are used to construct dynamic benchmarks.
Benchmark Self-Evolving~\citep{wang2024selfevolving} employs a multi-agent framework to dynamically extend existing static benchmarks, showcasing the potential of agent-based methods.
Given a task description, BENCHAGENTS~\citep{butt2024benchagents} leverages a multi-agent framework for automated benchmark creation. 
It splits the process into planning, generation, verification, and evaluation—each handled by a specialized LLM agent.
This coordinated method, with human-in-the-loop feedback, yields scalable, diverse, and high-quality benchmarks.

\fakeparagraph{Limitations}
The quality of LLM-generated samples is often uncertain. For instance, human annotation in LatestEval~\citep{li2023avoiding} reveals that ~10\% of samples lack faithfulness or answerability. In interactive settings, reliability further depends on the interviewer LLM.

\subsubsection{Hybrid Generation}
LatestEval~\citep{li2023avoiding} combines temporal cutoff and LLM-based generation to automatically generate reading comprehension datasets using LLMs on real-time content from sources such as BBC.
DARG~\citep{zhang2024darg} integrates LLM-based and graph-based generation. It first extracts reasoning graphs from existing benchmarks and then perturbs them into new samples using predefined rules.
C$^2$LEVA~\citep{li-etal-2025-c2leva} incorporates all three contamination-free construction methods to build a contamination-free bilingual evaluation.
TrustGen~\citep{huang2025trustworthiness} is the first dynamic benchmarking to evaluate trustworthiness across multiple dimensions and
model types, including text-to-image, large language, and vision-language models.

\section{Discussions}
\label{sec:discussion}


\fakeparagraph{Current Challenges}
Benchmarking LLMs is essential for evaluating model performance, but traditional static benchmarks risk data contamination. Dynamic benchmarks address this risk by updating or regenerating test data, aiming to maintain integrity. However, current dynamic methods often lack standardized evaluation criteria, suffer from limited scalability, and offer little interpretability. Many also fail to systematically assess trade-offs like computational overhead and robustness.


\fakeparagraph{Future Directions}
Future work should establish standardized evaluation frameworks with criteria such as correctness, diversity, and scalability. Contamination-resilient benchmarks—using temporal filtering, synthetic data, or rule-based generation—can further improve reliability. Dynamic benchmarks should also support continual updates, cross-model applicability, and human-in-the-loop validation. Public update logs and improved interpretability will enhance transparency and trust in LLM evaluation.
Future directions also include extending dynamic benchmarking to multi-modal LLMs~\citep{chen-etal-2024-beyond-single,chen2024voicebench}.

\vspace{-2mm}
\section{Conclusion}
\vspace{-2mm}
This survey reviews the literature on data contamination in LLM benchmarking, analyzing both static and dynamic methods. We find that static methods, though consistent, become more vulnerable to contamination as training datasets grow. While dynamic methods show promise, they face challenges in reliability and reproducibility. Future research should focus on standardized dynamic evaluation, and practical mitigation tools.

\section*{Acknowledgements}
Simin and Baishakhi are partially supported in part by CCF 2313055, CCF 2107405, CAREER 2025082, and FAI: 2040961. 
Dezhi and Tao are partially supported by National Natural Science Foundation of China under Grant No. 623B2006, and Grant No. 92464301. 
Any opinions, findings, conclusions, or recommendations expressed herein are those of the authors.

\section* {Limitations}

While this survey provides a comprehensive overview of static and dynamic benchmarking methods for LLMs, there are several limitations to consider. First, due to the rapidly evolving nature of LLM development and benchmarking methods, some recent methods or tools may not have been fully covered. As benchmarking practices are still emerging, the methods discussed may not yet account for all potential challenges or innovations in the field. Additionally, our proposed criteria for dynamic benchmarking are a first step and may need further refinement and validation in real-world applications. Finally, this survey focuses primarily on high-level concepts and may not delve into all the fine-grained technical details of specific methods, which may limit its applicability to practitioners seeking in-depth implementation guidelines.

\section* {Ethical Considerations}

Our work is rooted in the goal of enhancing the transparency and fairness of LLM evaluations, which can help mitigate the risks of bias and contamination in AI systems. However, ethical concerns arise when considering the use of both static and dynamic benchmarks. Static benchmarks, if not carefully constructed, can inadvertently perpetuate biases, especially if they rely on outdated or biased data sources. Dynamic benchmarks, while offering a more adaptive method, raise privacy and security concerns regarding the continual collection and updating of data. Moreover, transparency and the potential for misuse of benchmarking results, such as artificially inflating model performance or selecting biased evaluation criteria, must be carefully managed. It is essential that benchmarking frameworks are designed with fairness, accountability, and privacy in mind, ensuring that they do not inadvertently harm or disadvantage certain user groups or research domains. Finally, we encourage further exploration of ethical guidelines surrounding data usage, model transparency, and the broader societal impact of AI benchmarks.

\bibliography{anthology,custom}

\appendix

\balance

\begin{table*}[htbp]
  \centering
  
\resizebox{0.98\textwidth}{!}{
    \begin{tabular}{ccccccc}
    \toprule
    \toprule
    \multicolumn{1}{c}{\multirow{2}[2]{*}{Leakage}} & \multicolumn{3}{c}{HumanEval} & \multicolumn{3}{c}{DyCodeEval} \\
          & \multicolumn{1}{l}{Llama-3.2-1B} & \multicolumn{1}{l}{Llama-3.2-3B} & \multicolumn{1}{l}{DeepSeek-Coder-1.3b} & \multicolumn{1}{l}{Llama-3.2-1B} & \multicolumn{1}{l}{Llama-3.2-3B} & \multicolumn{1}{l}{DeepSeek-Coder-1.3b} \\
    \midrule
    0\%   & 0.19  & 0.28  & 0.41  & 0.14  & 0.25  & 0.41 \\
    25\%  & 0.29  & 0.32  & 0.47  & 0.08  & 0.18  & 0.13 \\
    50\%  & 0.48  & 0.57  & 0.50   & 0.08  & 0.19  & 0.16 \\
    75\%  & 0.68  & 0.71  & 0.59  & 0.07  & 0.21  & 0.14 \\
    100\% & 0.82  & 0.87  & 0.62  & 0.11  & 0.18  & 0.07 \\
    \bottomrule
    \bottomrule
    
    \end{tabular}%
    }
  \caption{A proof of concept experiment.}
  
\label{tab:contamination}%
\end{table*}%

\section{Significance of Data Contamination}
\label{sec:appendix_demo}
To demonstrate the effectiveness of dynamic benchmarks, we following existing work~\cite{chen2025dynamic} and conduct a study using HumanEval and DyCodeEval~\cite{chen2025dynamic} using three LLMs: Llama-3.2-1B, Llama-3.2-3B, and DeepSeek-Coder-1.3B. For each model, we simulate data contamination by intentionally leaking a portion of the benchmark dataset during fine-tuning.
We experiment with contamination levels of 0\%, 25\%, 50\%, 75\%, and 100\% respectively, producing four distinct contaminated models.

The results show that for overfitted models, as the contamination level increases from 25\% to 100\%, accuracy on HumanEval also increases. This  result highlights the limitation of static benchmarks in detecting overfitting. However, on the dynamic DyCodeEval, even when a model is overfitted on one version, it maintains stable accuracy scores across different versions. The results demonstrate the advantage of dynamic benchmarks in evaluating models under data contamination.

\begin{table*}[t!]
    \centering
    \small
    
    \begin{tabularx}{\textwidth}{llX}
    \toprule
    \toprule
    \textbf{Task} & \textbf{Type} & \textbf{Benchmark} \\
    \midrule
    \multirow{2}[2]{*}{\textbf{Math}} & Static & GSM8K~\citep{cobbe2021training}, MATH~\citep{hendrycks2021measuring}, AIME 2024~\citep{maa2024aime}, CNMO 2024)~\citep{cnmo2024} \\
          & Dynamic & LiveBench~\citep{white2024livebench}, UGMathBench~\citep{xu2025ugmathbench}, Mathador-LM~\citep{kurtic-etal-2024-mathador} \\
    \midrule
    \multirow{2}[2]{*}{\textbf{Language}} & Static & GLUE~\citep{wang-etal-2018-glue}, SuperGLUE~\citep{wang2019superglue}, CLUE~\citep{xu-etal-2020-clue} \\
          & Dynamic & LiveBench~\citep{white2024livebench}, C$^2$LEVA~\citep{li-etal-2025-c2leva}, ITD~\citep{zhu-etal-2024-inference} \\
    \midrule
    \multirow{2}[2]{*}{\textbf{Coding}} & Static & HumanEval~\citep{chen2021evaluating}, MBPP~\citep{austin2021program}, SWE-Bench~\citep{jimenez2024swebench,yang2024swebenchmultimodal}, Codeforces~\citep{codeforces}, Aider~\citep{aider} \\
          & Dynamic & PPM~\cite{chen2024ppm}, DyCodeEval~\cite{chen2025dynamic}, LiveCodeBench~\citep{jain2024livecodebenchholisticcontaminationfree}, ComplexCodeEval~\citep{10.1145/3691620.3695552} \\
    \midrule
    \multirow{2}[2]{*}{\textbf{Reasoning}} & Static & PIQA~\citep{bisk2020piqa}, SIQA~\citep{sap-etal-2019-social}, HellaSwag~\citep{zellers-etal-2019-hellaswag}, WinoGrande~\citep{sakaguchi2021winogrande}, ARC~\citep{clark2018think}, OpenBookQA~\citep{mihaylov-etal-2018-suit}, CommonsenseQA~\citep{talmor-etal-2019-commonsenseqa}, C-SimpleQA~\citep{he2024chinese} \\
          & Dynamic & LiveBench~\citep{white2024livebench}, DyVal~\citep{zhu2024dyval}, C$^2$LEVA~\citep{li-etal-2025-c2leva}, NPHardEval~\citep{fan-etal-2024-nphardeval}, S3Eval~\citep{lei-etal-2024-s3eval}, DARG~\citep{zhang2024darg} \\
    \midrule
    \multirow{2}[2]{*}{\textbf{Knowledge}} & Static & NaturalQuestions~\citep{kwiatkowski-etal-2019-natural}, TriviaQA~\citep{joshi-etal-2017-triviaqa}, CMMLU~\citep{li-etal-2024-cmmlu}, MMLU~\citep{hendrycks2020measuring}, BBH~\citep{suzgun-etal-2023-challenging}, AGI Eval~\citep{zhong2023agieval}, MMLU-Redux~\citep{gema-etal-2025-done}, MMLU-Pro~\citep{wang2024mmlu}, ControlBench~\citep{Darioush2024ControlBench}, FRAMES~\citep{krishna2024fact}, GPQA Diamond~\citep{rein2023gpqagraduatelevelgoogleproofqa}, AlpacaEval~\citep{alpaca_eval}, ArenaHard~\citep{arenahard2024} \\
          & Dynamic & C$^2$LEVA~\citep{li-etal-2025-c2leva}, ITD~\citep{zhu-etal-2024-inference}, Auto-Dataset~\citep{ying2024automating}, DyVal2~\citep{10.5555/3692070.3694661}, SciEval~\citep{sun2024scieval} \\
    \midrule
    \multirow{2}[2]{*}{\textbf{Safety}} & Static & RealToxicityPrompts~\citep{gehman-etal-2020-realtoxicityprompts}, ToxiGen~\citep{hartvigsen-etal-2022-toxigen} \\
          & Dynamic & C$^2$LEVA~\citep{li-etal-2025-c2leva}, FactBench~\citep{bayat2024factbench} \\
    \midrule
    \multirow{2}[2]{*}{\textbf{Instruction}} & Static & IFEval~\citep{zhou2023instruction}, InfoBench~\citep{qin-etal-2024-infobench}, C-Eval~\citep{huang2024c} \\
          & Dynamic & LiveBench~\citep{white2024livebench} \\
    \midrule
    \multirow{2}[2]{*}{\textbf{Comprehension}} & Static & SQuAD~\citep{rajpurkar-etal-2018-know}, QuAC~\citep{choi2018quac}, BoolQ~\citep{clark-etal-2019-boolq} \\
          & Dynamic & LatestEval~\citep{li2023avoiding}, Antileak-bench~\citep{wu2024antileak} \\
    \bottomrule
    \bottomrule
    \end{tabularx}
    \caption{Summary of benchmarking applications.}
    
    \label{tab:static-benchmarks}
\end{table*}

\section{Benchmark Applications}
\label{sec:appendix_applications}

\fakeparagraph{Knowledge}
Knowledge benchmarks evaluate LLM internal knowledge. NaturalQuestions~\citep{kwiatkowski-etal-2019-natural} and TriviaQA~\citep{joshi-etal-2017-triviaqa} focus on retrieving real-world information, while multi-domain tasks are covered by MMLU~\citep{hendrycks2020measuring}, BBH~\citep{suzgun-etal-2023-challenging}, and AGI Eval~\citep{zhong2023agieval}. Recent extensions like MMLU-Redux~\citep{gema-etal-2025-done} and MMLU-Pro~\citep{wang2024mmlu} refine these assessments further. 
Additionally, ControlBench~\citep{Darioush2024ControlBench}, FRAMES~\citep{krishna2024fact}, and GPQA Diamond~\citep{rein2023gpqagraduatelevelgoogleproofqa} target technical and long-context challenges, with open-domain evaluations provided by AlpacaEval~\citep{alpaca_eval} and ArenaHard~\citep{arenahard2024}.

\fakeparagraph{Reasoning}
Understanding and applying everyday knowledge is a key aspect of language comprehension. Benchmarks such as PIQA~\citep{bisk2020piqa}, SIQA~\citep{sap-etal-2019-social}, HellaSwag~\citep{zellers-etal-2019-hellaswag}, and WinoGrande~\citep{sakaguchi2021winogrande} are designed to assess a model’s intuitive reasoning skills from multiple perspectives. In addition, academic challenge sets like ARC~\citep{clark2018think}, OpenBookQA~\citep{mihaylov-etal-2018-suit}, and CommonsenseQA~\citep{talmor-etal-2019-commonsenseqa} push models further by requiring the integration of background knowledge with logical reasoning to arrive at plausible answers. C-SimpleQA~\citep{he2024chinese} evaluates the factuality ability of language models to answer short questions in Chinese.

\fakeparagraph{Safety}
Safety benchmarks are essential for evaluating the robustness of LLM’s ability to generate non-toxic and ethically aligned content. Datasets such as RealToxicityPrompts~\citep{gehman-etal-2020-realtoxicityprompts} and ToxiGen~\citep{hartvigsen-etal-2022-toxigen} assess resilience against producing harmful outputs. TrustGen~\citep{huang2025trustworthiness} is the first dynamic benchmarking to evaluate trustworthiness across multiple dimensions and
model types, including text-to-image, large language, and vision-language models.

\fakeparagraph{Language} Language benchmarks assess the LLMs’ proficiency in specific languages.
GLUE~\citep{wang-etal-2018-glue} and SuperGLUE~\citep{wang2019superglue} cover tasks from sentiment analysis to  language inference, while CLUE~\citep{xu-etal-2020-clue} targets Chinese language.
Typo-fixing~\citep{suzgun-etal-2023-challenging} is also widely used.

\fakeparagraph{Reading Comprehension}
Reading comprehension tasks test a model’s ability to extract and infer information from text. Benchmarks like SQuAD~\citep{rajpurkar-etal-2018-know}, QuAC~\citep{choi2018quac}, and BoolQ~\citep{clark-etal-2019-boolq} challenge models to understand passages and draw logical conclusions.

\section{Static Benchmark Enhancements}
\label{sec:enhancement}
Because LLMs often train on publicly available data, static benchmarks risk being inadvertently included, leading to contamination. To mitigate this risk, several methods have been proposed to enhance \textit{static} benchmarking.

\subsection{Canary String}
Canary strings are deliberately crafted, being unique tokens embedded within a dataset to serve as markers for data contamination. When a model’s output unexpectedly includes these tokens, it strongly indicates that the model has memorized portions of its training data rather than learning to generalize. For instance, the BIG-Bench dataset incorporates these strings so that model developers can identify and filter out such instances \cite{jacovi-etal-2023-stop}. 

\fakeparagraph{Limitations} The effectiveness of canary strings depends on model trainers being aware of and responsive to these markers. 
If a developer aims to leak benchmarking data to boost scores, this method will not work.

\subsection{Encryption}
Encryption methods secure evaluation data by making it inaccessible to unauthorized parties, preventing its accidental inclusion in training sets. \citet{jacovi-etal-2023-stop} propose encrypting test data with a public key and a ``No Derivatives’’ license to block automated crawling and reuse. \citet{yang2023rethinking} show that even advanced decontamination methods can be defeated by minor text variations, emphasizing the need for robust encryption. Similarly, TRUCE \citep{chandran2024private} leverages confidential computing and secure multi‐party computation to enable private benchmarking, ensuring that test data and model parameters remain confidential. 

\fakeparagraph{Limitations} While these methods effectively protect against data leakage, they depend on strong key management, they introduce extra computational overheads. These methods are vulnerable if encryption is compromised or private key is exposed.

\subsection{Label Protection}
Label protection involves keeping the true answers of a test set hidden from public access so that only an authorized evaluator can use them during model assessment. This method is common in benchmarks such as GLUE \cite{wang-etal-2018-glue}, SuperGLUE \cite{wang2019superglue}, and OpenAI’s HumanEval \cite{chen2021evaluating}.  where the test labels are withheld to prevent models from learning or memorizing them during training. The key advantage of this method is its ability to maintain evaluation integrity by preventing model exposure to answers, thereby mitigating data contamination risks.

\fakeparagraph{Limitations} Label protection limits transparency and independent verification, and it forces researchers to rely on centralized evaluation systems for performance metrics, which can impede detailed error analysis and reproducibility.

\subsection{Post-hoc Detection}
Post-hoc detection mitigates data contamination by identifying overlaps between $D_{train}$
  and $D_ 
{test}$. This method is typically done through n-gram matching at various levels, such as tokens~\cite{touvron2023llama} or words~\cite{radford2019language,brown2020language,chowdhery2023palm}. However, exact matching often leads to false negatives, prompting the use of more robust methods like embedding-based similarity~\cite{riddell-etal-2024-quantifying,lee2023platypus,gunasekar2023textbooks} and improved mapping metrics~\cite{li-etal-2024-open-source,xu2024benchmarking}.

Beyond direct overlap detection, post-hoc methods also analyze model behavior under different conditions, such as memorization through masked inputs~\cite{ranaldi-etal-2024-investigating,chang-etal-2023-speak}, partial completions~\cite{anil2023palm,golchin2024time}, or preference for original over paraphrased test cases~\cite{duartecop,golchin2023data,zong2024fool}. For instance, \citeauthor{dekoninck2024constat}~(\citeyear{dekoninck2024constat}) propose CONSTAT, which detects contamination by comparing model performance across benchmarks.

\fakeparagraph{Limitations}
Post-hot detection methods face several limitations. Full access to the training dataset is often restricted due to legal and privacy constraints, making overlap detection challenging. Additionally, assumptions about model behavior, such as higher memorization or lower perplexity for contaminated instances, may not hold across different models and tasks.

\section{Dynamic Benchmarking Strategy Property Labeling Guidance}
\label{sec:appendix_labeling}

We label each dynamic benchmark as ``supported'',
``partially supported'', or ``not supported'' for each criterion based on the following guidelines:

\fakeparagraph{Correctness} Benchmarks with built-in guarantees (e.g., via temporal cutoffs or rule-based generation) are marked ``supported''. LLM-generated benchmarks are ``partially supported'' if validated (e.g., by humans or automation), and ``not supported'' otherwise.

\fakeparagraph{Scalability}
Fully automated benchmarks are ``supported''. Those combining automation with human effort are ``partially supported'', while purely manual ones are ``not supported''.

\fakeparagraph{Collision}
If a benchmark provides theoretical guarantees or formally analyzes collision rates, it is ``supported''. Empirical analysis without guarantees is ``partial support'', and absence of discussion results in ``not supported''.

\fakeparagraph{Complexity Stability}
Benchmarks that define and control complexity are ``supported''. Those that define but do not control it receive ``partial support''. Lack of discussion results in ``not supported''.

\fakeparagraph{Diversity}
Benchmarks that define and enforce diversity are ``supported''. Those that define but do not control it are ``partially supported'', and benchmarks that omit it are ``not supported''.

\fakeparagraph{Interpretability} Rule-based or human-designed benchmarks are ``supported''. Those combining rules with LLMs receive ``partial support''. Benchmarks relying entirely on LLMs without interpretability are ``not supported''.

\end{document}